\documentclass[letterpaper, 10 pt, conference]{ieeeconf}  

\IEEEoverridecommandlockouts                              

\usepackage{url}            
\usepackage{booktabs}       
\usepackage{amsfonts}       
\usepackage{nicefrac}       
\usepackage{microtype}      
\usepackage{xcolor}         

\usepackage{blindtext}

\usepackage{mathtools}

\usepackage{url}

\usepackage{algorithm}
\usepackage{algorithmic}
\usepackage{verbatim}

\usepackage{mathtools}

\usepackage{microtype}
\usepackage{graphicx}
\usepackage{subfigure}
\usepackage{booktabs}

\usepackage[utf8]{inputenc} 
\usepackage[T1]{fontenc}    
\usepackage{hyperref}       
\usepackage{url}            
\usepackage{booktabs}       
\usepackage{amsfonts}       
\usepackage{nicefrac}       
\usepackage{microtype}      
\usepackage{xcolor}         
\usepackage{balance}
\usepackage{cite}

\title{\LARGE \bf
Learning Manipulation Skills through Robot Chain-of-Thought with Sparse Failure Guidance
}

\author{Kaifeng Zhang$^{1}$, Zhao-Heng Yin$^{2}$, Weirui Ye$^{3,1,4}$, Yang Gao$^{3,1,4}$ 
\thanks{$^1$Shanghai Qi Zhi Institute. $^2$UC Berkeley.  $^3$Tsinghua University. $^4$Shanghai Artificial Intelligence Laboratory.}
\thanks{Contact: {\tt\small zhangkf@lamda.nju.edu.cn}.}
}

\begin{document}

\maketitle
\thispagestyle{empty}
\pagestyle{empty}

\begin{abstract}

Defining reward functions for skill learning has been a long-standing challenge in robotics. Recently, vision-language models (VLMs) have shown promise in defining reward signals for teaching robots manipulation skills. However, existing work often provides reward guidance that is too coarse, leading to insufficient learning processes. In this paper, we address this issue by implementing more fine-grained reward guidance. We decompose tasks into simpler sub-tasks, using this decomposition to offer more informative reward guidance with VLMs. We also propose a VLM-based self imitation learning process to speed up learning. Empirical evidence demonstrates that our algorithm consistently outperforms baselines such as CLIP, LIV, and RoboCLIP. Specifically, our algorithm achieves a $5.4 \times$ higher average success rates compared to the best baseline, RoboCLIP, across a series of manipulation tasks.

\end{abstract}

\section{INTRODUCTION}\label{intro}

Defining reward functions for learning robotic skills has been a long-standing challenge in robotics~\cite{hussein2017imitation}. Traditional methods typically define rewards through human knowledge and require significant engineering efforts. Some methods also use expert demonstrations to define rewards~\cite{hussein2017imitation}. However, this procedure requires expert supervision and can be infeasible in complex scenarios~\cite{RoboCLIP} . Recently, the success of vision language models~(VLM) has inspired researchers to use its power to define rewards for skill learning. For example, RoboCLIP \cite{RoboCLIP} captures the temporal information from robot movements with an S3D video backbone \cite{S3D}, therefore defining the rewards for skill learning. 

In spite of the preliminary success of VLM-based reward modeling, a downside of these existing works is that their language guidance is too coarse for learning. Consider the task of opening the door, existing works only use the language embedding of "opening the door" in reward definition to guide the learning process. Such guidance can be too coarse for learning: When the robot is not able to approach the door at the beginning of learning process, the rewarder will always provide a near-zero feedback since the door is not likely to be opened. In contrast, in human world we typically use more fine-grained guidance such as "move towards the door", "move the hand towards the handle" and "grab the handle" during the teaching process. Intuitively, using such decomposition \cite{CoT, HiP, SayCan, ViLa, LP} can make the guidance more precise, and will lead to richer and denser feedback. 

Based on this intuition, in this paper, we present a novel framework named Rewarder from \textbf{Robo}t \textbf{C}hain-\textbf{o}f-\textbf{T}hought with Sparse Failure Guidance (RoboCoT) to address this problem. Our main insight is to decompose the task description into a series of actionable steps and then leverage VLM to define a fine-grained reward to guide skill learning, which can make the learning process more sample efficient and yield higher success rates. In addition, we propose a VLM-based self-imitation procedure to further speed up the learning process. In the experiments, we demonstrate that our approach performs well across a broad spectrum of tasks, achieving a $5.4\times$ performance improvement over the best existing baseline. Empirical studies show that each component of our algorithm contributes to the overall performance. 

\begin{figure*}[t]
\begin{center} 
\includegraphics[width=6.5in]{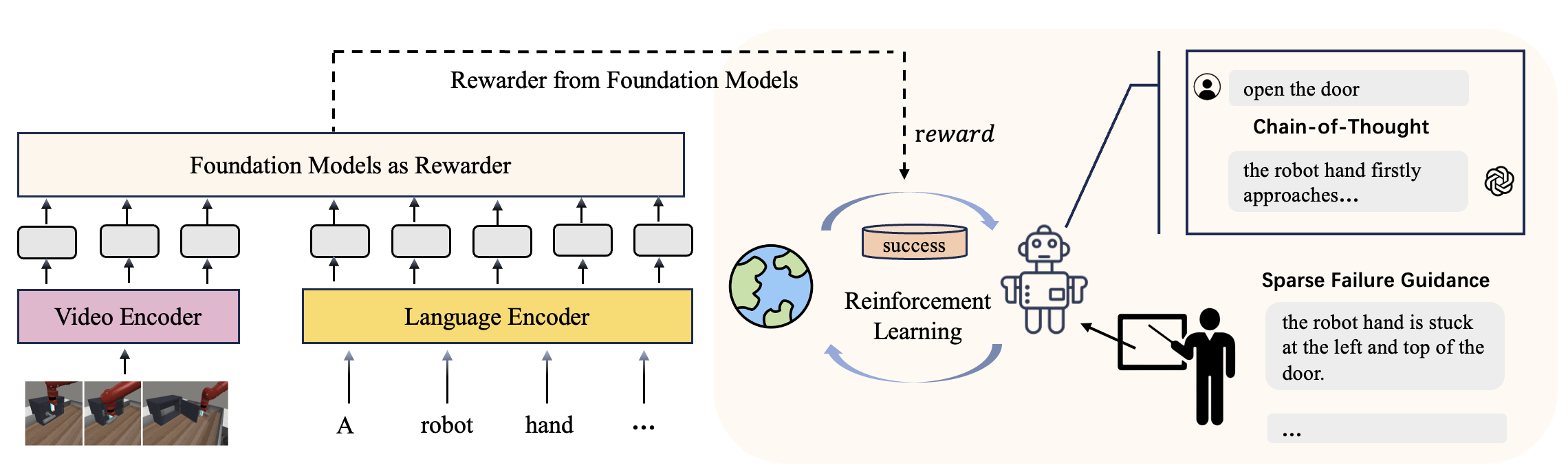} 
\caption{First, the robot receives the language instruction, e.g., "open the door." Subsequently, our algorithm interprets this instruction using a robot chain-of-thought processing, breaking it down into several detailed prompts. Then, the robot initiates reinforcement learning with our designed reward signal to learn the skills with the consideration about the failure guidance. Successful experiences are recorded by foundation models to reinforce effective behaviors, i.e., VLM-based self-imitation.}
\label{concept} 
\end{center} 
\end{figure*}

\section{RELATED WORK}

Defining reward functions for robotic skill learning has long been a significant challenge. This challenge can be addressed through learning from demonstrations using inverse reinforcement learning (IRL) \cite{GAIL, liu_imitation, jeong2020self, matas2018sim, xu2023roboninja}. However, IRL requires expert supervision, which can be infeasible in complex scenarios \cite{RoboCLIP}. Our RoboCoT overcomes this limitation by effectively leveraging VLMs as a source of rewards. By fully utilizing prior knowledge from VLMs, RoboCoT comprehends the manipulation process and subsequently achieves superior manipulation performance by learning directly from language instructions.

\subsection{Robot Chain-of-Thought for Robotics}

Robot chain-of-thought \cite{CoT} is employed to interpret task descriptions and enhance high-level planning performance in robotics. For example, ViLa \cite{ViLa} utilizes GPT-4V to decompose tasks into sub-goals based on robot observations and task descriptions, demonstrating significant success in understanding complex robotic tasks. HiP \cite{HiP} achieves hierarchical planning by diffusing the future observations, therefore better decomposing the complex tasks. However, current approaches primarily apply robot chain-of-thought to high-level planning, thereby limiting its potential for comprehensively understanding robotic tasks. Our RoboCoT extends the application of robot chain-of-thought by designing more fine-grained reward signals. Empirical studies demonstrate its promise in directly following language instructions.

\subsection{Vision Language Models for Robotics}

VLMs equip robot agents with compact, pre-trained representations. R3M \cite{R3M} co-trains video and language instructions in a self-supervised manner \cite{TCN}, enhancing the robot's ability to mimic expert actions. PaLM-E \cite{PaLME} integrates visual observations with large language models, facilitating low-level robot control that includes reasoning about the control process. RT-1 \cite{RT1} uses transformers to better mimic expert behaviors. RoboFlamingo \cite{RoboFlamingo} uses multi-step observation inputs to capture a comprehensive view of observational data. RT-2 \cite{RT2} innovates by conceptualizing robot actions as a form of language. RT-H \cite{RTH} introduces an action hierarchy through language motions, leveraging the robot chain-of-thought reasoning capabilities of large language models (LLMs).

VLMs significantly enhance robot manipulation capabilities by offering valuable reward signals. CLIP \cite{CLIP}, for instance, aligns image representations with language embeddings. Consequently, \cite{MetaCLIP} employs CLIP to measure the reward distance between a robot's current observation and its linguistic goal. LIV \cite{LIV} innovatively combines VIP \cite{VIP} and CLIP to more effectively learn both visual and linguistic goal representations. RoboCLIP \cite{RoboCLIP} adopts the S3D \cite{S3D} representation model to assess the reward for the final observation of a trajectory. Additionally, \cite{FG} considers potential failure cases for designing reward signals in imitating language instructions. RL-VLM-F \cite{RL_VLM_F} directly uses GPT-4V to label preferences for agent rollout data.

However, current models use coarse instructions like "open the door" for designing reward signals. Our RoboCoT improves upon this by leveraging the robot chain-of-thought technique to design more fine-grained reward signals for following language instructions. Empirical studies demonstrate the promise of using VLMs as rewarders in imitation learning.

\section{BACKGROUND}

We consider scenarios where the robotic system receives specific verbal commands, such as "open the door." To model this learning process, we employ a Markov Decision Process (MDP) defined by $(S, A, r, P, \gamma)$, where $S$ denotes the state space, $A$ the action space, $r$ the reward model, $P$ the transition model, and $\gamma$ the discount factor.

\textbf{Chain-of-Thought (CoT)}
The CoT \cite{CoT} approach is a reasoning method designed to enhance the problem-solving capabilities of LLMs. This method involves decomposing a problem into a series of intermediate steps or thoughts that collectively guide towards a solution, mirroring the human approach to complex problem solving. By explicitly detailing these intermediate steps, LLMs such as GPT-4V are better equipped to navigate and resolve complex challenges that require advanced understanding and logical deduction.

\textbf{VideoCLIP} 
VideoCLIP \cite{VideoCLIP} is a multi-modal model that connects visual and textual information, specifically designed to understand videos by associating them with relevant text descriptions. The model is trained on a diverse dataset of videos and associated texts, allowing it to develop a nuanced understanding of the relationship between video segments and language. In particular, VideoCLIP minimizes the sum of two multi-modal contrastive losses:
\begin{align}
    \mathcal{L} = - \sum_{(v,t) \in \mathcal{B}}(\log \text{NCE}(z_v,z_t)+\log \text{NCE}(z_t,z_v)),
\end{align}
where $\mathcal{B}$ is the batch containing sampled video-text pairs, and NCE$(z_v, z_t)$ and NCE$(z_t, z_v)$ correspond to the contrastive loss on video-to-text similarity and vice versa. Specifically, the video-to-text contrastive loss is given by:
\begin{align}
    \text{NCE}(z_v,z_t) = \frac{\exp(z_v \cdot z_t^+/\tau)}{\sum_{z\in{\{z_t^+, z_t^-\}}}\exp(z_v \cdot z/\tau)},
\end{align}
where $\tau$ is a temperature hyper-parameter, $z_t^+$ and $z_t^-$ are positive and negative text embeddings respectively, and $z_v$ is a video segment embedding.

\section{METHOD}
\begin{figure*}
\begin{center} 
\includegraphics[width=6.5in]{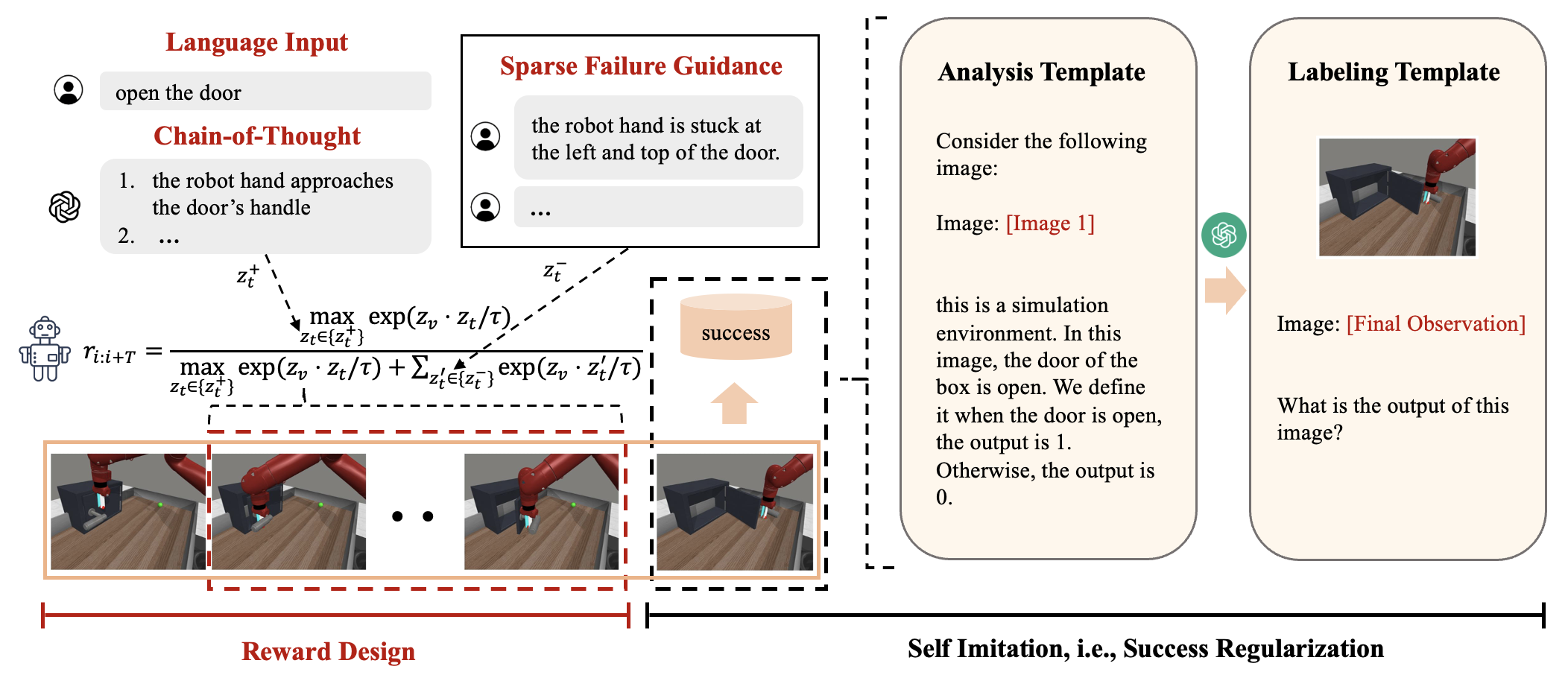} 
\caption{First, text embeddings are generated with robot chain-of-thought (positive prompts) and sparse failure guidance (negative prompts). Next, a moving window slides across the agent's rollout trajectory, calculating the NCE reward based on both video and text embeddings. Finally, VLMs evaluate the final observation. If this observation is deemed successful, this trajectory is recorded in the replay buffer for self imitation procedure.}
\label{method} 
\end{center} 
\end{figure*}

In Section \ref{prompt_gen}, we present the robot chain-of-thought process for comprehending the robotic tasks. In Section \ref{rewarder}, we outline the process of defining rewards for skill learning using Vision-Language Models (VLMs). Finally, in Section \ref{success_reg}, we introduce the VLM-based self-imitation procedure through success experience regularization.

\subsection{Robot Chain-of-Thought}\label{prompt_gen}

Robotic manipulation skills consist of a series of actionable control steps. For example, opening a door involves sub-tasks such as approaching the handle, grasping it, pulling it back, and opening the door. Decomposing tasks with a robot chain-of-thought (CoT) process simplifies the manipulation process, providing robots with a deeper understanding and greater control.

However, previous works \cite{RoboCLIP, FG, MetaCLIP, LIV} have primarily used coarse language instructions, such as 'open the door,' in defining reward signals. This level of guidance is insufficient for learning robust manipulation skills. Our RoboCoT addresses this limitation by employing robot chain-of-thought process to decompose tasks for more fine-grained reward design.

We use GPT-4V to interpret task instructions like 'open the door.' Through robot chain-of-thought process, GPT-4V breaks down this instruction into a series of sub-goals, such as 'the robot hand approaches the door's handle,' 'the robot hand grasps the door's handle,' 'the robot hand pulls the door's handle back,' and 'the robot hand opens the door.' Consequently, we leverage these task descriptions from the GPT-4V interpretation to define a fine-grained reward signal in skill learning, thereby enhancing the agent's understanding of the robotic tasks.

\subsection{VLMs as Rewarder}\label{rewarder}

Defining reward functions for robot skill learning is a long standing challenge. In this work, we leverage VLMs to design more fine-grained reward signals in skill learning. Moreover, we also consider the potential failure guidance in teaching robot manipulation skills which has been investigated in \cite{FG}. 

We explore the usage of VideoCLIP \cite{VideoCLIP} as a rewarder. VideoCLIP excels in capturing the temporal dynamics of video slices and interpreting associated language prompts, aiming at aligning robot movements with actions described in the task descriptions. 

We define the reward signal for skill learning as:
\begin{align}\label{reward}
    r = \frac{\max \limits_{z_l \in \{z_l^+\}} \exp(z_v \cdot z_l/\tau)}{\max \limits_{z_l \in \{z_l^+\}}\exp(z_v \cdot z_l/\tau) + \sum \limits_{z'_l\in{\{z_l^-\}}}\exp(z_v \cdot z'_l/\tau)},
\end{align}
where $r$ represents the rewards for moving window stride timesteps, $z_v$ is the video embedding obtained using a moving window sliding on the agent rollout trajectory, $z_l^+$ is a set of detailed task description embeddings from robot chain-of-thought process, and $z_l^-$ is a set of potential failure case embeddings guided by humans. Here, sparse failure guidance motivates the agent to consider potential failure cases, improving the robustness. 

The training of VideoCLIP, employing a methodology akin to the InfoNCE loss, aims at distinguishing between video slices and corresponding positive or negative texts. We find that this similar NCE-based reward formulation can better leverage the VideoCLIP model which is different from the reward definition in RoboCLIP \cite{RoboCLIP}. Moreover, Equation \ref{reward} ensures that robot behaviors aligning with the detailed task descriptions and avoiding failure scenarios yield higher rewards to the agent.

\begin{figure*}
\begin{center} 
\vskip 0.2in
\includegraphics[width=6.5in]{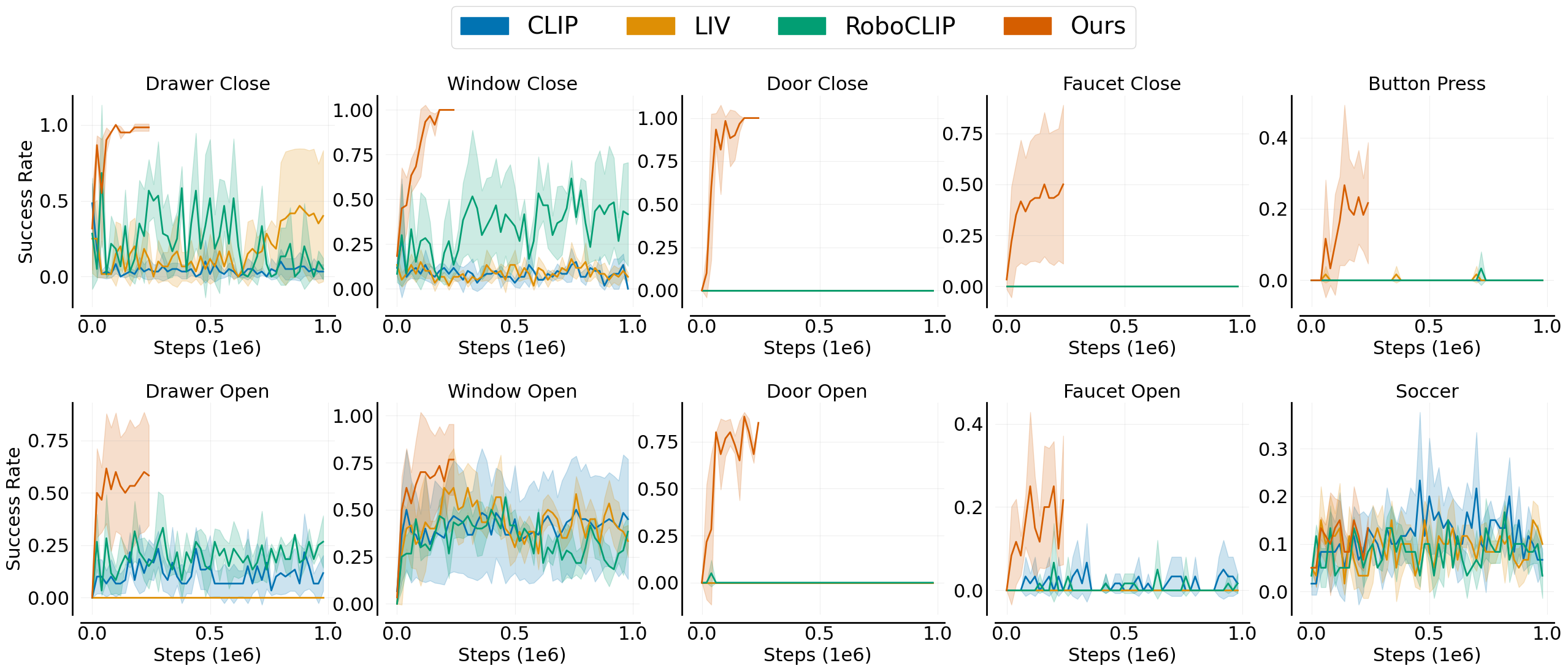} 
\caption{Training curves for baselines and our algorithm within 3 random seeds. Each data point is evaluated with 20 sampled trajectories. The shaded area displays the range of one standard deviation. Our algorithm gets best performance compared to other baselines, achieves $5.4\times$ average improvement compared to RoboCLIP.}
\label{main} 
\end{center} 
\end{figure*}

\subsection{VLM-based Self Imitation}\label{success_reg}

Our reward design mechanism encourages the agent to explore the necessary skills. We find that the VideoCLIP rewards focus on whether the robot movements align with the detailed language instructions. Consequently, speeding up the learning is necessary. To better speed up the learning, we introduce VLM-based self imitation procedure. 

We employ VLMs, i.e., GPT-4V, to assess the success of the agent's final observation in its rollout trajectory. If it deemed successful, this trajectory is recorded into a success replay buffer $D_E$. Additionally, we correct the reward function on this successful observation as: $r(o_T) = r(o_T) + 100.0$. Here, $r(o_T)$ is the reward function for this successful final observation $o_T$ in the rollout trajectory. Subsequently, we introduce a regularization term in loss function, aimed at reinforcing success behaviors. The regularization loss is integrated with the policy RL loss, formulated as follows:
\begin{align}\label{policy_loss}
    \mathcal{L} = \text{RL Loss} + \lambda \sum_{(s,a) \in D_E}-\log(\pi(a|s)),
\end{align}
where $D_E$ is the success experience replay buffer and $\lambda$ is the hyper-parameter. By incorporating success experiences into the policy loss calculation, we speed up the skill learning process. Here we note that in our experiments, $\lambda$ is set to $1$.

\section{EXPERIMENT}

In this section, we assess the effectiveness of our proposed algorithm RoboCoT. The experiments are designed to address two questions: (1) How does our algorithm RoboCoT perform compared to other state-of-the-art algorithms? The results are presented in section \ref{overall}. (2) What are the contributing factors in our RoboCoT? The ablation studies are presented in section \ref{ablation_study}.

\textbf{Environments} To evaluate the effectiveness of our proposed algorithm RoboCoT, we conduct experiments on a wide array of manipulation tasks using MetaWorld-v2 as the test robotic environment. We select 10 tasks from simple to complex, including drawer, window, faucet, button, door, and soccer, as described in \cite{ROT}. Among these tasks, opening objects such as drawers, windows, faucets, and doors is significantly more challenging than closing them.

\textbf{Setup} To ensure a fair evaluation across all experiments, we use DrQ-v2 \cite{DrQv2} as the reinforcement learning algorithm. Detailed implementation specifics can be found in the references: CLIP \cite{MetaCLIP}, LIV \cite{LIV}, and RoboCLIP \cite{RoboCLIP}. All the experiments, including those involving our proposed algorithm RoboCoT, receive the same language instructions as inputs.

\subsection{Overall Comparison}\label{overall}

To demonstrate the superiority of our algorithm, we conduct a series of experiments comparing our RoboCoT with existing baselines: CLIP \cite{MetaCLIP}, LIV \cite{LIV}, and RoboCLIP \cite{RoboCLIP}. 

CLIP \cite{MetaCLIP} employs CLIP \cite{CLIP} as the VLMs to measure the distance between current observations and language goals. LIV \cite{LIV} aligns VIP \cite{VIP} goal representations with language instructions, therefore measuring the reward distance between current robot observation with language goals. RoboCLIP \cite{RoboCLIP} utilizes S3D \cite{S3D} to measure the distance between trajectory videos and language as a form of reward. Since the reward is defined with a coarse language instruction, CLIP \cite{CLIP}, LIV \cite{LIV} and RoboCLIP \cite{RoboCLIP} lack fine-grained reward guidance. On the other hand, our RoboCoT uses robot chain-of-thought and sparse failure guidance to define the rewards for skill learning. Consequently, RoboCoT can better acquire the skills through reinforcement learning.

Figure \ref{main} illustrates the training curves, depicting the success rates for each algorithm. Each evaluation involves 20 trajectories, and the experiments were replicated across three different environmental random seeds to introduce variability in the position of objects.

\begin{figure*}
\vskip 0.2in
\includegraphics[width=6.5in]{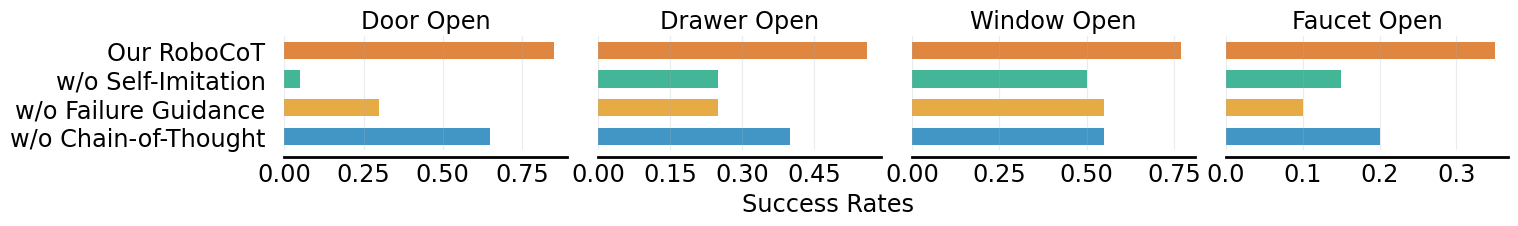} 
\vskip 0.5mm
\includegraphics[width=6.5in]{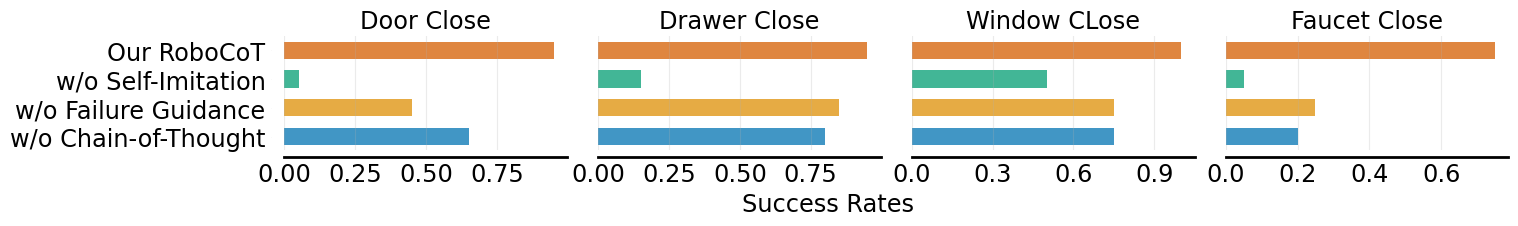} 
\vskip 0.5mm
\includegraphics[width=3.8in]{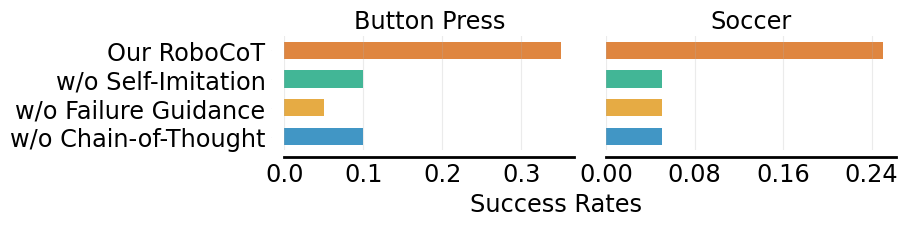} 
\caption{Final policy performance for ablations and our RoboCoT. The results show that each component of our RoboCoT contributes to the final performance.}
\label{ablation} 

\end{figure*}

Our findings indicate distinct performance trends across various tasks. RoboCLIP displays promising results in the simplest tasks. CLIP performs well in tasks related to faucet, drawer and window manipulation. Because CLIP focuses more on task completion rather than comprehending the robotic manipulation process. LIV shows competence primarily in tasks involving obvious task goal definition. Our RoboCoT consistently outperforms the baselines across all tasks, achieving success rates exceeding $75\%$ in tasks such as drawers, windows, and doors. These results underscore the effectiveness of our RoboCoT in handling diverse and challenging scenarios.

\subsection{Ablation Study}\label{ablation_study}

In this section, we conduct ablation studies by removing each component of our algorithm including without self-imitation procedure, without sparse failure guidance and without robot Chain-of-Thought. The results indicate that each component of our RoboCoT contributes to the overall performance. In addition, we conduct ablation studies comparing GPT-4V with ground truth labels in collecting success experience, demonstrating the robustness of GPT-4V labeling. Finally, experiments with all baselines incorporating VLM-based self imitation procedure reveal that our algorithm's reward design mechanism is superior compared to other baselines.

\emph{\textbf{Question 1: how significant is each component of our algorithm in contributing to the overall performance?}}

Our algorithm incorporates several critical components: VLM-based self imitation procedure, sparse human failure guidance, and robot Chain-of-Thought process to define the rewards. To evaluate the significance of each component, we conducted a series of experiments, systematically removing each component to observe its impact on performance. A decline in performance upon the removal of any component demonstrates its essential role in our RoboCoT.

\begin{figure*}
\begin{center} 

\includegraphics[width=6.5in]{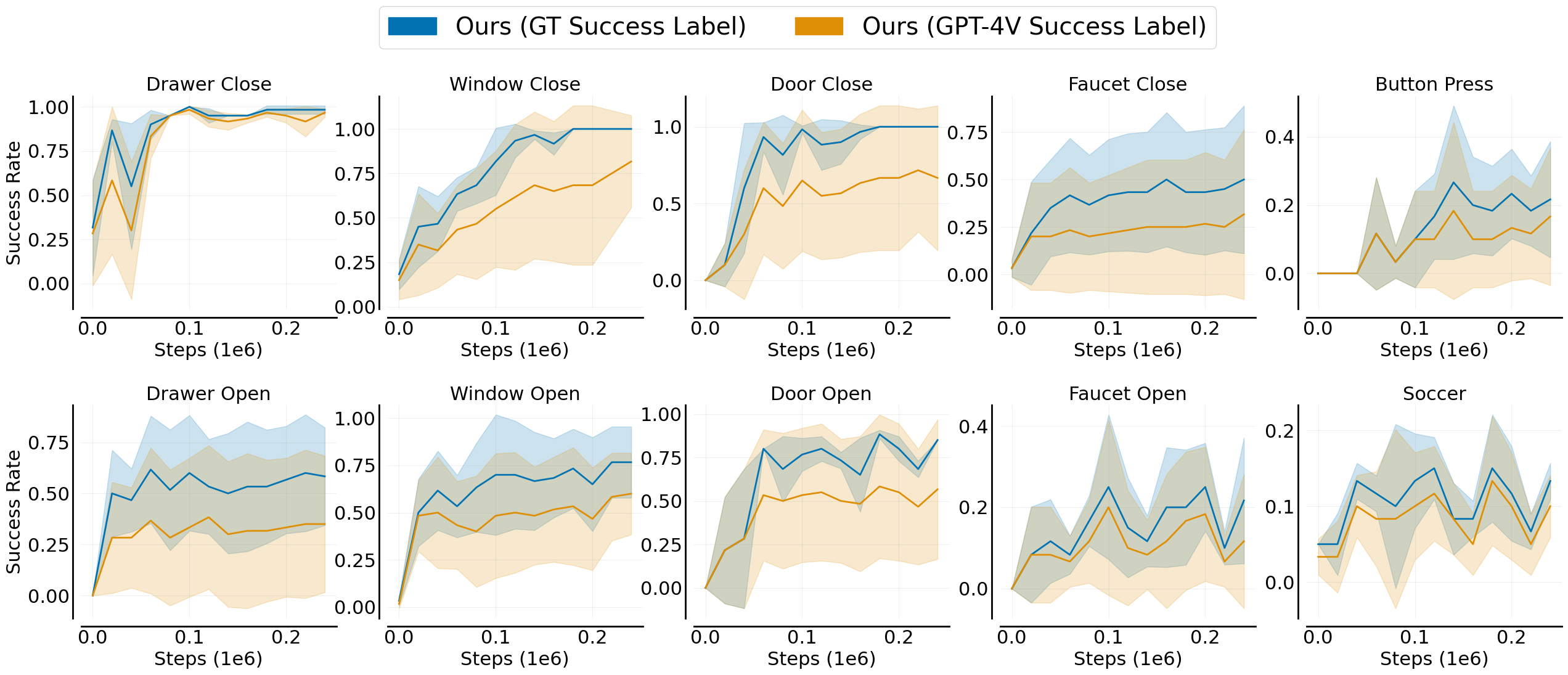} 
\caption{Ablation study for our algorithm with different success experience collection methods (GPT-4V or ground truth (GT) labels). The results demonstrate that the GPT-4V are robust in collecting success experience in a subset of robotic tasks.}
\label{gpt} 
\end{center} 
\end{figure*}

\begin{figure*}
\begin{center} 
\includegraphics[width=6.5in]{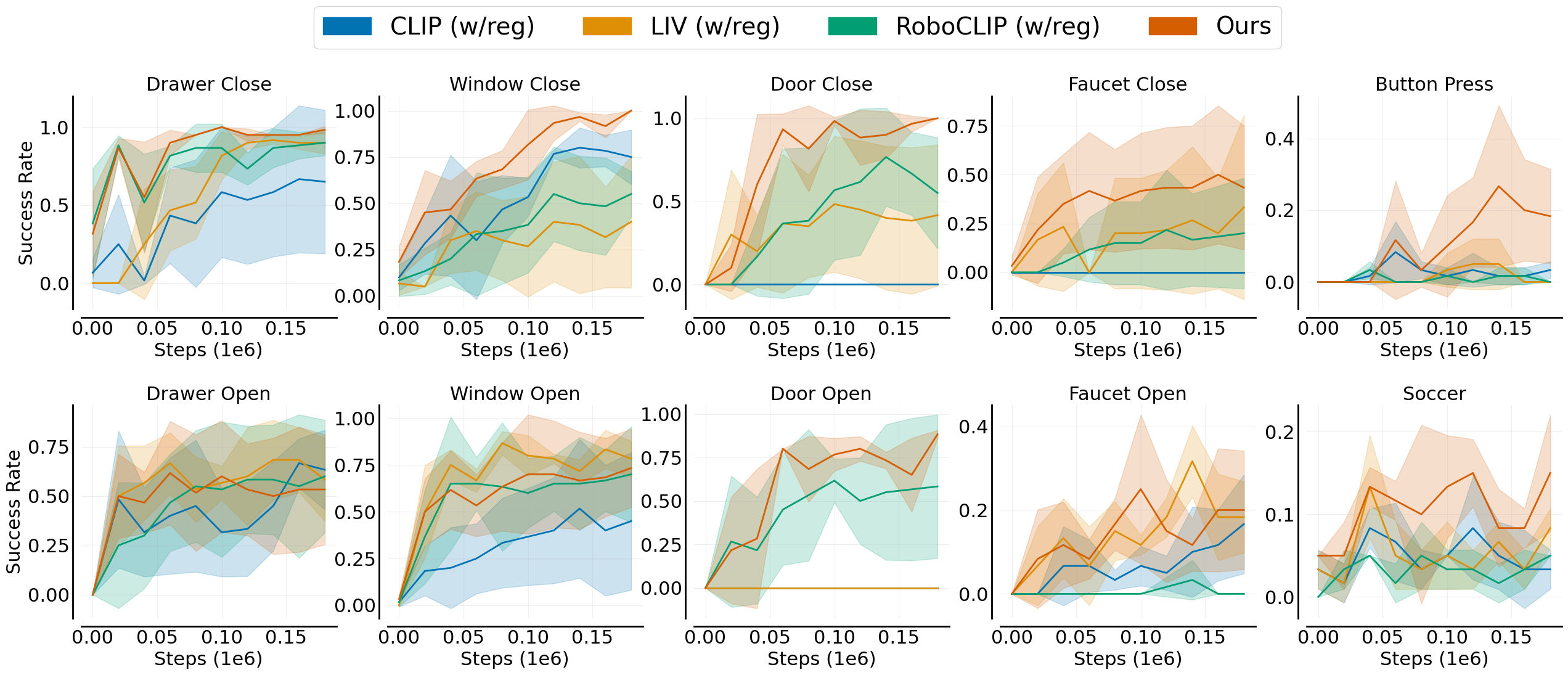} 
\caption{Ablation study for all baselines with VLM-based self imitation procedure (w/reg). The results show that the reward design mechanism in our RoboCoT is superior than other baselines.}
\label{wbc} 

\end{center} 
\end{figure*}

Figure \ref{ablation} illustrates the success rates of the final policy performance of our RoboCoT compared to its ablated versions. The labels for these ablations are without self-imitation procedure, without failure guidance and without Chain-of-Thought process.  

The results confirm the critical importance of each component in our algorithm. Each component in our RoboCoT contributes to the overall performance. In particular, VLM-based self imitation plays an important role in our RoboCoT. For better evaluating the importance of reward design mechanism in RoboCoT, we also conduct an ablation study with all baselines incorporating VLM-based self imitation procedure in Question 3.

\emph{\textbf{Question 2: Is GPT-4V robust in the context of success experience collection?}}

Our algorithm RoboCoT incorporates GPT-4V to collect success experience for VLM-based self imitation procedure. To validate the robustness of our GPT-4V labels in collecting success experience, we conduct experiments comparing GPT-4V labels to ground truth labels. 

The process for GPT-4V analysis is outlined in Figure \ref{method}. Initially, we present GPT-4V with a scenario indicating a successful state (e.g., the door is open). We then provide a prompt: "Consider the following image; this is a simulation environment. In this image, the door of the box is open. We define it when the door is open, the output is 1. Otherwise, the output is 0." Subsequently, we present the final observation to GPT-4V and prompt, "What is the output of this image?" We collect the trajectory as successful when GPT-4V's response is 1.

Figure \ref{gpt} presents the training success rate curves comparing GPT-4V and ground truth (GT) labels for collecting success experience. Our findings indicate that GPT-4V's labeling performance declines in part of the tasks. However, in most tasks, the performance is comparable to that of the ground truth labels, demonstrating that GPT-4V labels are robust for a subset of robotic tasks.

\emph{\textbf{Question 3: Is the reward design mechanism superior in our RoboCoT compared to other baselines?}}

Question 1 shows that VLM-based self imitation component plays an important role in our RoboCoT. To evaluate the significance of the reward design mechanism, we conduct experiments incorporating VLM-based self imitation procedure to all baselines. If the results show that our algorithm still outperforms other baselines with VLM based self imitation procedure, then it proves that our algorithm's reward design mechanism is superior than other baselines.

Figure \ref{wbc} shows the empirical results comparing our RoboCoT with other state-of-the-art baselines which also incorporating VLM-based self imitation procedure. The results demonstrate that our algorithm outperforms other baselines on almost all tasks. It indicates that the reward design mechanism in our RoboCoT is superior in skill learning compared to other existing baselines.

\section{CONCLUSIONS}

In this study, we present a novel skill learning algorithm, i.e., RoboCoT, which provides more fine-grained reward guidance for learning manipulation skills. Our RoboCoT simplifies complex tasks through a robot chain-of-thought process. It motivates the robot to better comprehend the complex tasks for skill learning. Through VLM-based self imitation procedure, our RoboCoT speeds up the learning process, thereby improving its performance. Empirical results confirm the effectiveness of our approach, demonstrating a distinct advantage compared to other state-of-the-art baselines. This research sets a new standard for mastering manipulation skills in robotics. We hope our RoboCoT can be applied into more complex robotic tasks in the future.




\bibliographystyle{IEEEtran}
\balance
\bibliography{root}

\begin{thebibliography}{10}
\providecommand{\url}[1]{#1}
\csname url@rmstyle\endcsname
\providecommand{\newblock}{\relax}
\providecommand{\bibinfo}[2]{#2}
\providecommand\BIBentrySTDinterwordspacing{\spaceskip=0pt\relax}
\providecommand\BIBentryALTinterwordstretchfactor{4}
\providecommand\BIBentryALTinterwordspacing{\spaceskip=\fontdimen2\font plus
\BIBentryALTinterwordstretchfactor\fontdimen3\font minus \fontdimen4\font\relax}
\providecommand\BIBforeignlanguage[2]{{%
\expandafter\ifx\csname l@#1\endcsname\relax
\typeout{** WARNING: IEEEtran.bst: No hyphenation pattern has been}%
\typeout{** loaded for the language `#1'. Using the pattern for}%
\typeout{** the default language instead.}%
\else
\language=\csname l@#1\endcsname
\fi
#2}}

\bibitem{hussein2017imitation}
A.~Hussein, M.~M. Gaber, E.~Elyan, and C.~Jayne, ``Imitation learning: A survey of learning methods,'' \emph{ACM Computing Surveys (CSUR)}, vol.~50, no.~2, pp. 1--35, 2017.

\bibitem{RoboCLIP}
\BIBentryALTinterwordspacing
S.~A. Sontakke, J.~Zhang, S.~Arnold, K.~Pertsch, E.~Biyik, D.~Sadigh, C.~Finn, and L.~Itti, ``Robo{CLIP}: One demonstration is enough to learn robot policies,'' in \emph{Thirty-seventh Conference on Neural Information Processing Systems (NeurIPS)}, 2023. [Online]. Available: \url{https://openreview.net/forum?id=DVlawv2rSI}
\BIBentrySTDinterwordspacing

\bibitem{S3D}
S.~Xie, C.~Sun, J.~Huang, Z.~Tu, and K.~Murphy, ``Rethinking spatiotemporal feature learning: Speed-accuracy trade-offs in video classification,'' in \emph{Proceedings of the European conference on computer vision (ECCV)}, 2018, pp. 305--321.

\bibitem{CoT}
J.~Wei, X.~Wang, D.~Schuurmans, M.~Bosma, F.~Xia, E.~Chi, Q.~V. Le, D.~Zhou, \emph{et~al.}, ``Chain-of-thought prompting elicits reasoning in large language models,'' \emph{Advances in neural information processing systems (NeurIPS)}, vol.~35, pp. 24\,824--24\,837, 2022.

\bibitem{HiP}
\BIBentryALTinterwordspacing
A.~Ajay, S.~Han, Y.~Du, S.~Li, A.~Gupta, T.~S. Jaakkola, J.~B. Tenenbaum, L.~P. Kaelbling, A.~Srivastava, and P.~Agrawal, ``Compositional foundation models for hierarchical planning,'' in \emph{Thirty-seventh Conference on Neural Information Processing Systems (NeurIPS)}, 2023. [Online]. Available: \url{https://openreview.net/forum?id=dyXNh5HLq3}
\BIBentrySTDinterwordspacing

\bibitem{SayCan}
M.~Ahn, A.~Brohan, N.~Brown, Y.~Chebotar, O.~Cortes, B.~David, C.~Finn, C.~Fu, K.~Gopalakrishnan, K.~Hausman, \emph{et~al.}, ``Do as i can, not as i say: Grounding language in robotic affordances,'' \emph{arXiv preprint arXiv:2204.01691}, 2022.

\bibitem{ViLa}
Y.~Hu, F.~Lin, T.~Zhang, L.~Yi, and Y.~Gao, ``Look before you leap: Unveiling the power of gpt-4v in robotic vision-language planning,'' \emph{arXiv preprint arXiv:2311.17842}, 2023.

\bibitem{LP}
W.~Huang, P.~Abbeel, D.~Pathak, and I.~Mordatch, ``Language models as zero-shot planners: Extracting actionable knowledge for embodied agents,'' \emph{arXiv preprint arXiv:2201.07207}, 2022.

\bibitem{GAIL}
J.~Ho and S.~Ermon, ``Generative adversarial imitation learning,'' \emph{Advances in neural information processing systems (NeurIPS)}, vol.~29, 2016.

\bibitem{liu_imitation}
Y.~Liu, W.~Dong, Y.~Hu, C.~Wen, Z.-H. Yin, C.~Zhang, and Y.~Gao, ``Imitation learning from observation with automatic discount scheduling,'' \emph{International Conference on Learning Representations (ICLR)}, 2024.

\bibitem{jeong2020self}
R.~Jeong, Y.~Aytar, D.~Khosid, Y.~Zhou, J.~Kay, T.~Lampe, K.~Bousmalis, and F.~Nori, ``Self-supervised sim-to-real adaptation for visual robotic manipulation,'' in \emph{2020 IEEE international conference on robotics and automation (ICRA)}.\hskip 1em plus 0.5em minus 0.4em\relax IEEE, 2020, pp. 2718--2724.

\bibitem{matas2018sim}
J.~Matas, S.~James, and A.~J. Davison, ``Sim-to-real reinforcement learning for deformable object manipulation,'' in \emph{Conference on Robot Learning (CoRL)}.\hskip 1em plus 0.5em minus 0.4em\relax PMLR, 2018, pp. 734--743.

\bibitem{xu2023roboninja}
Z.~Xu, Z.~Xian, X.~Lin, C.~Chi, Z.~Huang, C.~Gan, and S.~Song, ``Roboninja: Learning an adaptive cutting policy for multi-material objects,'' \emph{arXiv preprint arXiv:2302.11553}, 2023.

\bibitem{R3M}
S.~Nair, A.~Rajeswaran, V.~Kumar, C.~Finn, and A.~Gupta, ``R3m: A universal visual representation for robot manipulation,'' \emph{Conference on Robot Learning (CoRL)}, 2022.

\bibitem{TCN}
P.~Sermanet, C.~Lynch, Y.~Chebotar, J.~Hsu, E.~Jang, S.~Schaal, S.~Levine, and G.~Brain, ``Time-contrastive networks: Self-supervised learning from video,'' in \emph{2018 IEEE international conference on robotics and automation (ICRA)}.\hskip 1em plus 0.5em minus 0.4em\relax IEEE, 2018, pp. 1134--1141.

\bibitem{PaLME}
D.~Driess, F.~Xia, M.~S. Sajjadi, C.~Lynch, A.~Chowdhery, B.~Ichter, A.~Wahid, J.~Tompson, Q.~Vuong, T.~Yu, \emph{et~al.}, ``Palm-e: An embodied multimodal language model,'' \emph{arXiv preprint arXiv:2303.03378}, 2023.

\bibitem{RT1}
A.~Brohan, N.~Brown, J.~Carbajal, Y.~Chebotar, J.~Dabis, C.~Finn, K.~Gopalakrishnan, K.~Hausman, A.~Herzog, J.~Hsu, \emph{et~al.}, ``Rt-1: Robotics transformer for real-world control at scale,'' \emph{arXiv preprint arXiv:2212.06817}, 2022.

\bibitem{RoboFlamingo}
X.~Li, M.~Liu, H.~Zhang, C.~Yu, J.~Xu, H.~Wu, C.~Cheang, Y.~Jing, W.~Zhang, H.~Liu, \emph{et~al.}, ``Vision-language foundation models as effective robot imitators,'' \emph{International Conference on Learning Representations (ICLR)}, 2024.

\bibitem{RT2}
A.~Brohan, N.~Brown, J.~Carbajal, Y.~Chebotar, X.~Chen, K.~Choromanski, T.~Ding, D.~Driess, A.~Dubey, C.~Finn, \emph{et~al.}, ``Rt-2: Vision-language-action models transfer web knowledge to robotic control,'' \emph{arXiv preprint arXiv:2307.15818}, 2023.

\bibitem{RTH}
S.~Belkhale, T.~Ding, T.~Xiao, P.~Sermanet, Q.~Vuong, J.~Tompson, Y.~Chebotar, D.~Dwibedi, and D.~Sadigh, ``Rt-h: Action hierarchies using language,'' \emph{arXiv preprint arXiv:2403.01823}, 2024.

\bibitem{CLIP}
A.~Radford, J.~W. Kim, C.~Hallacy, A.~Ramesh, G.~Goh, S.~Agarwal, G.~Sastry, A.~Askell, P.~Mishkin, J.~Clark, \emph{et~al.}, ``Learning transferable visual models from natural language supervision,'' in \emph{International conference on machine learning (ICML)}.\hskip 1em plus 0.5em minus 0.4em\relax PMLR, 2021, pp. 8748--8763.

\bibitem{MetaCLIP}
J.~Rocamonde, V.~Montesinos, E.~Nava, E.~Perez, and D.~Lindner, ``Vision-language models are zero-shot reward models for reinforcement learning,'' \emph{arXiv preprint arXiv:2310.12921}, 2023.

\bibitem{LIV}
Y.~J. Ma, V.~Kumar, A.~Zhang, O.~Bastani, and D.~Jayaraman, ``Liv: Language-image representations and rewards for robotic control,'' in \emph{International Conference on Machine Learning (ICML)}.\hskip 1em plus 0.5em minus 0.4em\relax PMLR, 2023, pp. 23\,301--23\,320.

\bibitem{VIP}
Y.~J. Ma, S.~Sodhani, D.~Jayaraman, O.~Bastani, V.~Kumar, and A.~Zhang, ``Vip: Towards universal visual reward and representation via value-implicit pre-training,'' \emph{International Conference on Leanring Representations (ICLR)}, 2023.

\bibitem{FG}
\BIBentryALTinterwordspacing
K.~Baumli, S.~Baveja, F.~Behbahani, H.~Chan, G.~Comanici, S.~Flennerhag, M.~Gazeau, K.~Holsheimer, D.~Horgan, M.~Laskin, C.~Lyle, H.~Masoom, K.~McKinney, V.~Mnih, A.~Neitz, D.~Nikulin, F.~Pardo, J.~Parker-Holder, J.~Quan, T.~Rocktäschel, H.~Sahni, T.~Schaul, Y.~Schroecker, S.~Spencer, R.~Steigerwald, L.~Wang, and L.~Zhang, ``Vision-language models as a source of rewards,'' 2023. [Online]. Available: \url{https://arxiv.org/abs/2312.09187}
\BIBentrySTDinterwordspacing

\bibitem{RL_VLM_F}
Y.~Wang, Z.~Sun, J.~Zhang, Z.~Xian, E.~Biyik, D.~Held, and Z.~Erickson, ``Rl-vlm-f: Reinforcement learning from vision language foundation model feedback,'' \emph{arXiv preprint arXiv:2402.03681}, 2024.

\bibitem{VideoCLIP}
H.~Xu, G.~Ghosh, P.-Y. Huang, D.~Okhonko, A.~Aghajanyan, F.~Metze, L.~Zettlemoyer, and C.~Feichtenhofer, ``Videoclip: Contrastive pre-training for zero-shot video-text understanding,'' \emph{Conference on Empirical Methods in Natural Language Processing (EMNLP)}, 2021.

\bibitem{ROT}
S.~Haldar, V.~Mathur, D.~Yarats, and L.~Pinto, ``Watch and match: Supercharging imitation with regularized optimal transport,'' in \emph{Conference on Robot Learning}.\hskip 1em plus 0.5em minus 0.4em\relax PMLR, 2023, pp. 32--43.

\bibitem{DrQv2}
D.~Yarats, R.~Fergus, A.~Lazaric, and L.~Pinto, ``Mastering visual continuous control: Improved data-augmented reinforcement learning,'' \emph{arXiv preprint arXiv:2107.09645}, 2021.

\end{thebibliography}
\addtolength{\textheight}{-12cm}





\end{document}